\documentclass[journal]{IEEEtran}
\IEEEoverridecommandlockouts

\usepackage{cite}
\usepackage{amsmath,amssymb,amsfonts}
\usepackage{algorithmic}
\usepackage{graphicx}
\usepackage{textcomp}
\usepackage{xcolor}
\usepackage{booktabs}
\usepackage{multirow}
\usepackage[colorlinks,linkcolor=red,anchorcolor=red,citecolor=blue]{hyperref}
\usepackage[lined,boxed,ruled]{algorithm2e}

\graphicspath{{./figs/}}

\def\BibTeX{{\rm B\kern-.05em{\sc i\kern-.025em b}\kern-.08em
    T\kern-.1667em\lower.7ex\hbox{E}\kern-.125emX}}
\begin{document}

\title{Minimal MMAO: A Resource-Closed-Loop Framework for Adaptive Metaheuristic Search}

\author{Jinliang Xu$^{*}$ and Liping Ma%
\thanks{Jinliang Xu received the bachelor's and master's degrees from Beijing University of Posts and Telecommunications, Beijing, China, where he also received the Ph.D. degree in computer science in 2020. He is with the China Academy of Information and Communications Technology, Beijing, China. His current research interests include heuristic optimization, agent-based applications, and blockchain systems. His e-mail: jlxufly@gmail.com and xujinliang@caict.ac.cn.}%
\thanks{Liping Ma is with the Department of Disease Control and Prevention, The Seventh Medical Center of Chinese PLA General Hospital, Beijing, China; e-mail: lipingmaqzx@163.com.}}

\maketitle

\begin{abstract}
This paper presents the Metabolic Multi-Agent Optimizer (MMAO) as an adaptive metaheuristic built around endogenous resource circulation. The central premise is that search intensity, exploration--exploitation balance, and lifecycle turnover should be induced by a shared metabolic controller rather than by separately attached schedules. We formulate MMAO through bounded private energy, a communal budget, normalized reward, continuous role adaptation, and resource-financed branching and pruning. The method is then instantiated in both continuous and discrete domains and evaluated on a matched small-scale suite including Sphere, Rastrigin, a synthetic Euclidean TSP, and two TSPLIB instances. The results show a consistent pattern: the same metabolic loop remains workable across domains, the discrete realization remains relatively stable under a compact design, and continuous refinement quality is the main cost of keeping the method lean. These results support MMAO's framework formulation under simplification, with some loss of late-stage precision on harder continuous tasks.
\end{abstract}

\begin{IEEEkeywords}
Adaptive metaheuristics, endogenous resource allocation, exploration--exploitation control, multi-agent optimization, parameter-light optimization.
\end{IEEEkeywords}

\section{Introduction}
Modern heuristic search increasingly relies on adaptive components such as dynamic populations, online operator selection, self-adjusting step scales, and budget reallocation across subpopulations or tasks. These developments are useful, but they also raise a recurring methodological problem. As algorithms become more adaptive, the organizing principle behind that adaptation often becomes less explicit. It then becomes difficult to tell whether a method is a coherent framework or only a successful bundle of mechanisms \cite{eiben1999parameter, karafotias2015parameter, li2014adaptivebandit, dong2025effective}.

This paper focuses on a different organizing principle, namely \emph{metabolic control}. Instead of prescribing separate schedules for search radius, exploitation intensity, population size, and reinvestment, MMAO lets these behaviors emerge from a resource-closed loop shared by all agents. Each agent maintains bounded private energy, successful search contributes to a communal pool, normalized reward converts problem-dependent progress into comparable signals, and lifecycle events are paid for from the same budget that sustains search. The shared budget regulates how search effort is spent and replenished.

The central question of this paper is whether such a formulation can remain informative and operational when written in a compact, explicit form. The objective is not to maximize benchmark coverage, but to expose the framework identity clearly enough that later theoretical analyses, domain derivations, or larger validation studies can build on it without ambiguity. We therefore emphasize three goals:
\begin{enumerate}
    \item formulate the essential MMAO state variables and resource flows in a domain-agnostic way;
    \item give continuous and discrete instantiations driven by the same metabolic loop;
    \item evaluate the empirical behavior of the framework under matched tasks and budgets, so that its strengths and limitations are visible rather than implied.
\end{enumerate}

This contraction study tests whether MMAO retains its resource-coupled adaptive behavior after simplification.

The main contributions of this paper are as follows.
\begin{enumerate}
    \item We formalize MMAO as a compact adaptive metaheuristic framework based on five coupled principles: bounded private energy, a communal pool, normalized reward, continuous role dynamics, and resource-financed lifecycle turnover.
    \item We provide domain-specific realizations for continuous and permutation search while preserving a single shared control logic.
    \item We present a mechanism mapping that distinguishes framework-essential components from optional instantiation detail.
    \item We report a matched small-scale experimental study across four continuous tasks and three TSP tasks, including TSPLIB \texttt{eil51} and \texttt{berlin52}, together with budget-fairness explanations and representative convergence traces.
\end{enumerate}

\section{Related Work}
MMAO is most closely related to research on endogenous parameter control, resource allocation, and adaptive population management. In evolutionary computation, the distinction between exogenous parameter schedules and endogenous self-adaptation has long been emphasized \cite{eiben1999parameter, karafotias2015parameter, brest2006self, Hansen2001}. More recent work has broadened this perspective toward dynamic population sizes, budget redistribution, and online learning of search behavior \cite{Tanabe2014, li2022distributed, liu2022cooperative, doerr2025speeding, cho2025configx, zhang2025laos}.

From the continuous-optimization side, MMAO also connects to derivative-free search and self-adaptive refinement. Symmetric probing and progress-normalized local sensing resemble a lightweight zero-order view of search, while continuous role modulation provides an endogenous alternative to explicit mode switching \cite{nesterov2017random, stripinis2024benchmarking, omeradzic2024self}. From the discrete side, the route-construction and route-improvement instantiations place MMAO near structure-aware TSP heuristics, though the present goal is not to compete with highly specialized methods such as Lin--Kernighan variants or strong neural-improvement pipelines \cite{Voudouris1999, Helsgaun2017, heins2025repair, ye2023deepaco, verdu2025scaling}.

What distinguishes MMAO from these lines of work is not any single operator. Rather, it is the insistence that adaptation should be derived from a closed metabolic budget. Search, reinforcement, and turnover are not treated as loosely coupled modules; they are consequences of one common resource economy. This study examines that shared resource-control structure.

\section{MMAO Framework Formulation}
\subsection{State and Resource Loop}
Let $\mathcal{P}_t=\{A_1,\dots,A_{N_t}\}$ denote the active population at iteration $t$. Each agent carries the state
\begin{equation}
    A_i(t)=\big(x_i(t), E_i(t), \phi_i(t), m_i(t)\big),
\end{equation}
where $x_i(t)$ is the current candidate solution, $E_i(t)\in[0,E_{\max}]$ is bounded private energy, $\phi_i(t)\in[0,1]$ is a continuous role variable, and $m_i(t)$ is local memory such as a personal best or structural elite route.

At the population level, the framework maintains a communal budget $B_t$ and a progress scale $s_t$. Raw improvement $\Delta_i(t)$ is mapped into normalized gain
\begin{equation}
    g_i(t)=\frac{\max(0,\Delta_i(t))}{s_t+\varepsilon},
\end{equation}
so that reward reacts to relative progress rather than to the absolute magnitude of the objective function.

The metabolic loop is then expressed abstractly as
\begin{equation}
    E_i(t+1)=\Pi_{[0,E_{\max}]}\left(E_i(t)+(1-\alpha)R(g_i(t))-C_i(t)\right),
\end{equation}
\begin{equation}
    B_{t+1}=\max\left(0,B_t+\alpha R(g_i(t))-U_t\right),
\end{equation}
where $R(\cdot)$ is a bounded reward transform, $\alpha$ is the communal share, $C_i(t)$ is private maintenance, and $U_t$ is the resource spent on branching, pruning, or respawn. In this formulation, adaptation is not attached externally. It is induced by changes in gain, energy, role, and communal budget.

\subsection{Framework Principles}
The present study treats the following five principles as framework-defining.
\begin{itemize}
    \item \textbf{Bounded private energy:} each agent has a finite local budget and cannot search indefinitely at full intensity.
    \item \textbf{Communal pool:} part of successful progress is socialized and can later be reinvested elsewhere in the population.
    \item \textbf{Normalized reward:} adaptation depends on scale-aware progress signals rather than raw objective differences.
    \item \textbf{Continuous role dynamics:} agents move on an exploration--exploitation continuum rather than belonging to hard-coded species.
    \item \textbf{Lifecycle turnover:} branching, pruning, and respawn are paid for by the same resource loop that governs search itself.
\end{itemize}

\begin{figure}[t]
\centering
\includegraphics[width=\columnwidth]{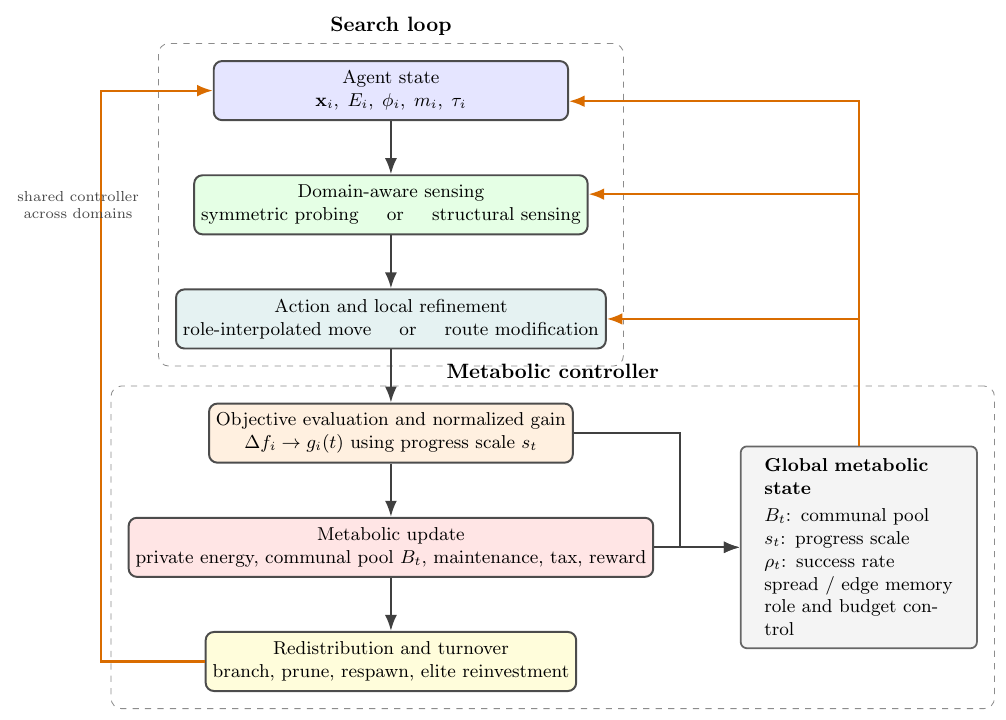}
\caption{Resource-closed-loop architecture of MMAO. Search behavior, reward, and lifecycle turnover are all regulated by the same metabolic controller.}
\label{fig:architecture}
\end{figure}

These principles are illustrated in Fig.~\ref{fig:architecture}, which shows the closed interaction between state, sensing, action, evaluation, metabolism, and turnover.

\section{Continuous and Discrete Instantiations}
\subsection{Continuous Instantiation}
In the continuous setting, each agent performs symmetric local sensing around its current state, forms a role-conditioned candidate move by combining probe information with global and personal guidance, and then updates its energy and role according to normalized gain. Agents with sustained gains can finance local offspring around promising anchors, while depleted agents are pruned and the communal pool may support respawn. The implementation therefore combines zero-order sensing, lightweight memory, and lifecycle control without introducing a separate external scheduler.

\subsection{Discrete Instantiation}
In the permutation setting, the same loop is realized through sampled 2-opt improvement, route kick/mutation, and lightweight edge-memory reconstruction. Role values regulate the balance between local route exploitation and more disruptive route modification, while the communal budget controls whether high-performing routes seed new candidates. This design aims to preserve structural awareness without collapsing into a task-specific handcrafted TSP solver.

\subsection{Mechanism Mapping}
To clarify what belongs to the framework and what belongs only to a particular realization, Table~\ref{tab:mapping} organizes MMAO by control responsibility rather than by an unstructured list of operators. The main point is that the framework is defined by the metabolic loop and its control obligations, not by any one motion equation or route operator.

This distinction is especially important in a contraction paper. The goal here is not to optimize every realization detail, but to test how much of the MMAO identity survives after the mechanism set is deliberately compressed. If the loop remains coherent after simplification, bounded energy, shared budget, role drift, and turnover provide a compact account of its adaptive behavior.

\begin{table*}[t]
\caption{Layered mechanism view of MMAO. Framework responsibilities are separated from domain-specific realizations.}
\label{tab:mapping}
\centering
\scriptsize
\setlength{\tabcolsep}{5pt}
\begin{tabular}{p{2.2cm}p{3.0cm}p{4.5cm}p{4.5cm}}
\toprule
\textbf{Layer} & \textbf{Control item} & \textbf{Framework responsibility} & \textbf{Realization in this paper} \\
\midrule
Metabolic state & Private energy & Bounded local effort and survival state & Kept as agent-level energy reserve \\
Metabolic state & Communal pool & Shared reinvestment budget across agents & Kept as population-level resource account \\
Metabolic state & Normalized reward & Makes progress comparable across scales & Kept through rolling gain normalization \\
Adaptive behavior & Continuous role state & Interpolates exploration and exploitation & Kept as a continuous role variable \\
Adaptive behavior & Lifecycle turnover & Finances branch, prune, and respawn from the same loop & Kept as budget-driven turnover \\
Continuous realization & Symmetric probing & Provides local zero-order sensing & Used in compact probe-based form \\
Continuous realization & Personal/global coupling & Blends memory and population guidance & Used in compact role-conditioned moves \\
Continuous realization & Elite precision refinement & Extends late-stage search sharpness & Not emphasized in the current compact setup \\
Discrete realization & Sampled 2-opt improvement & Provides route-level local refinement & Used directly \\
Discrete realization & Route kick / mutation & Supports escape from route stagnation & Used directly \\
Discrete realization & Edge memory bias & Reuses shared structural information & Used in lightweight form \\
Discrete realization & Extra reinvestment heuristics & Strengthens route reconstruction detail & Not emphasized in the current compact setup \\
\bottomrule
\end{tabular}
\end{table*}

\section{Algorithmic Procedure}
Algorithm~\ref{alg:mmao} summarizes the common MMAO loop. Domain-specific proposal and evaluation routines differ, but the metabolic controller is shared.

\begin{algorithm}[t]
\caption{Minimal MMAO Framework}
\label{alg:mmao}
\KwIn{Objective or cost function $f$, search domain $\Omega$, initial population $\mathcal{P}_0$, initial communal budget $B_0$, iteration limit $T$}
\KwOut{Best solution found $x^\star$}
Initialize agent states $(x_i,E_i,\phi_i,m_i)$ and communal budget $B_0$\;
Estimate initial progress scale $s_0$\;
\For{$t=0$ \KwTo $T-1$}{
    \ForEach{agent $A_i \in \mathcal{P}_t$}{
        Perform domain-aware sensing around $x_i(t)$\;
        Generate one or more role-conditioned candidate states\;
        Evaluate candidates and compute raw improvement $\Delta_i(t)$\;
        Convert $\Delta_i(t)$ into normalized gain $g_i(t)$\;
        Update private energy $E_i(t+1)$ and communal budget $B_{t+1}$\;
        Update role state $\phi_i(t+1)$ from gain and energy pressure\;
        Refresh local memory $m_i(t+1)$ and global best if improved\;
    }
    Prune agents with exhausted energy if population is above the minimum size\;
    Reinvest communal resources into branching or respawn when budget and parent quality permit\;
    Update progress scale $s_t$ from recent positive gains\;
}
Return the best solution encountered\;
\end{algorithm}

\begin{figure}[t]
\centering
\includegraphics[width=\columnwidth]{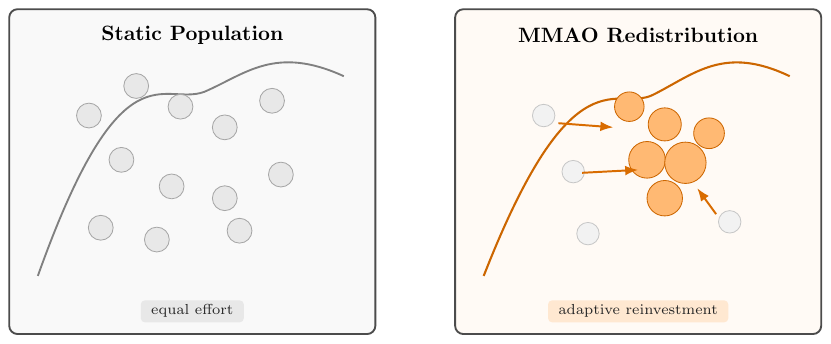}
\caption{Conceptual contrast between a static equal-effort population and MMAO-style adaptive redistribution.}
\label{fig:comparison}
\end{figure}

Figure~\ref{fig:comparison} provides an intuitive view of why the framework differs from a static equal-effort population: resources are allowed to concentrate around productive regions, but only through endogenous gain and budget feedback.

\section{Experimental Protocol}
\subsection{Tasks and Evaluation}
The purpose of the experiments is to test whether the framework behaves coherently and meaningfully in both continuous and discrete domains (see GitHub repository \texttt{mmao}\footnote{\url{https://github.com/wolfbrother/mmao}} or PyPI package \texttt{mmao-opt}\footnote{\url{https://pypi.org/project/mmao-opt/}}). We therefore use a matched small-scale suite rather than a large benchmark campaign.

The continuous side contains Sphere-2D, Rastrigin-2D, Sphere-5D, and Rastrigin-5D. These tasks provide a progression from smooth unimodal landscapes to more multimodal and dimensionally demanding settings. The discrete side contains a synthetic 10-city Euclidean TSP together with TSPLIB \texttt{eil51} and \texttt{berlin52}. For TSPLIB instances, we report gap to the known optimum. For the synthetic 10-city instance, we compute the exact optimum by Held--Karp dynamic programming.

\subsection{Budget Fairness}
The goal of the study is not to claim superiority over unrelated optimizers, but to assess the behavior of the MMAO framework under a consistent resource envelope. Fairness is therefore enforced in three ways.
\begin{itemize}
    \item Both continuous and discrete instantiations use the same five seeds $\{3,7,11,19,23\}$ on every task.
    \item For each task, the compared variants use the same iteration budget and the same minimum/initial/maximum population limits.
    \item The only intended difference is the level of mechanism detail inside the same metabolic controller, not the external search budget.
\end{itemize}

This design should be interpreted as \emph{within-framework fairness}. It supports diagnosis of the framework behavior itself, not broad benchmarking against unrelated algorithm families.

\section{Experimental Results}
\subsection{Summary Tables}
Tables~\ref{tab:contraction-cont} and~\ref{tab:contraction-tsp} summarize the results. On the continuous tasks, MMAO remains functional in all settings, but the more compact realization gives up a visible amount of final precision. The effect is mild on easy Sphere cases and becomes much stronger on Rastrigin-5D, where the mean best fitness rises from $1.7915$ to $8.0783$. This indicates that the common metabolic loop is sufficient for stable behavior, but rich late-stage refinement remains important when the landscape becomes more multimodal.

On the discrete tasks, the picture is more stable. Both variants solve the synthetic 10-city problem optimally. On \texttt{eil51} and \texttt{berlin52}, the compact realization incurs moderate gap increases, yet the degradation is far smaller than on the difficult continuous case. This suggests that the resource-closed-loop idea transfers well to structure-aware permutation search even when the realization remains relatively lean.

\begin{table}[t]
\caption{Continuous-task summary. Lower is better.}
\label{tab:contraction-cont}
\centering
\small
\resizebox{\columnwidth}{!}{%
\begin{tabular}{lcccc}
\toprule
\textbf{Task} & \textbf{Iter} & \textbf{Mean} & \textbf{Std} & \textbf{Best} \\
\midrule
Sphere-2D (MMAO) & 120 & 0.000001 & 0.000001 & 0.000000 \\
Sphere-2D (compact) & 120 & 0.006778 & 0.005201 & 0.001184 \\
Rastrigin-2D (MMAO) & 140 & 0.199374 & 0.398056 & 0.000180 \\
Rastrigin-2D (compact) & 140 & 0.400422 & 0.313415 & 0.110167 \\
Sphere-5D (MMAO) & 180 & 0.000003 & 0.000000 & 0.000002 \\
Sphere-5D (compact) & 180 & 0.024599 & 0.013085 & 0.010674 \\
Rastrigin-5D (MMAO) & 220 & 1.791463 & 0.744584 & 0.995365 \\
Rastrigin-5D (compact) & 220 & 8.078335 & 0.971423 & 6.775656 \\
\bottomrule
\end{tabular}
\vspace{1pt}}
\end{table}

\begin{table}[t]
\caption{TSP summary. Lower gap is better.}
\label{tab:contraction-tsp}
\centering
\small
\resizebox{\columnwidth}{!}{%
\begin{tabular}{lcccc}
\toprule
\textbf{Task} & \textbf{Iter} & \textbf{Mean Gap (\%) MMAO} & \textbf{Mean Gap (\%) compact} & \textbf{Best compact} \\
\midrule
Synthetic-10 & 160 & 0.00 & 0.00 & 27.9274 \\
\texttt{eil51} & 260 & 3.05 & 4.84 & 433 \\
\texttt{berlin52} & 280 & 4.58 & 6.08 & 7547 \\
\bottomrule
\end{tabular}}
\end{table}

\subsection{Representative Convergence Behavior}
\begin{figure}[t]
\centering
\includegraphics[width=\columnwidth]{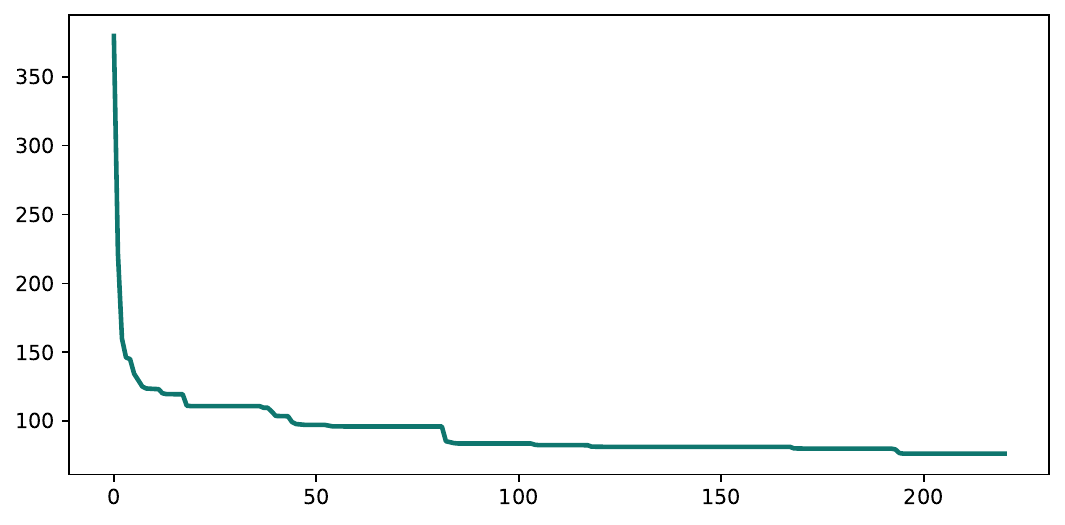}
\vspace{3pt}
\includegraphics[width=\columnwidth]{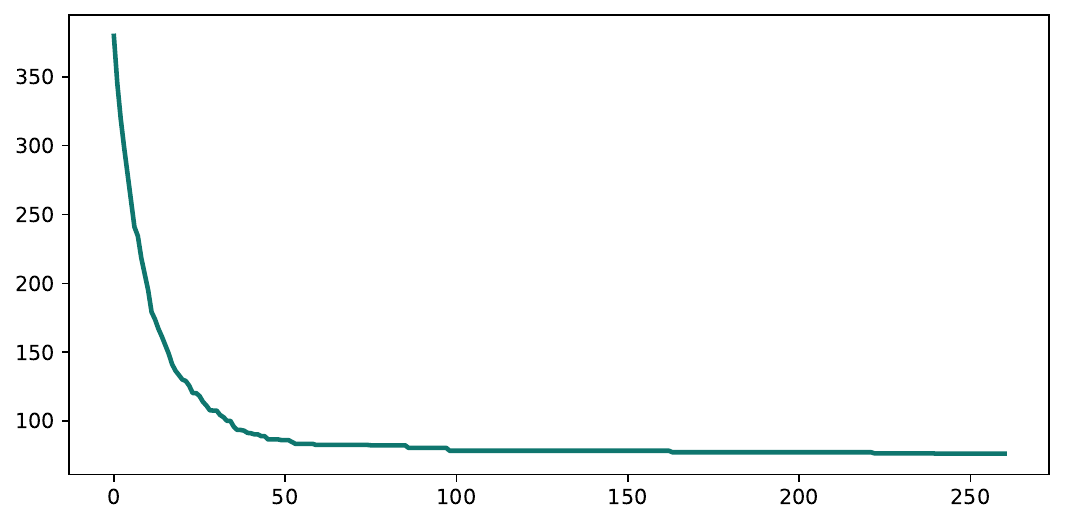}
\caption{Representative mean convergence traces on Rastrigin-5D (top) and \texttt{eil51} (bottom).}
\label{fig:conv}
\end{figure}

Representative convergence traces on Rastrigin-5D and \texttt{eil51} are shown in Fig.~\ref{fig:conv}. The continuous trace highlights that the richer instantiation keeps improving later into the search, whereas the compact variant stabilizes earlier. By contrast, the discrete trace shows a much smaller separation between the two curves, which is consistent with the gap statistics in Table~\ref{tab:contraction-tsp}.

\subsection{Extended Analysis}
The continuous and discrete sides expose different consequences of using the same closed-loop control principle. On the continuous side, the main challenge is not merely to find improvement, but to preserve sufficiently fine-grained adaptation once the population has already concentrated. This is why the difference is small in terms of qualitative convergence, yet large in terms of final precision. On Sphere, both variants enter the attraction basin reliably, but the compact realization already loses several orders of magnitude of terminal accuracy. On Rastrigin, especially in 5D, the gap widens further because multimodality makes later reinvestment quality and proposal sharpness much more important.

The discrete side is more forgiving. Once route-level local improvement, occasional disruptive kicks, and shared edge-memory bias are all present, the same metabolic controller can redistribute effort effectively without needing many additional task-specific mechanisms. This is most obvious in the synthetic 10-city case, where the compact realization preserves exact-quality behavior. It remains true, though to a lesser extent, on \texttt{eil51} and \texttt{berlin52}. The compact version degrades, but the relative increase in mean objective remains modest compared with the continuous degradation on Rastrigin-5D.

The different outcomes across domains suggest that the same resource economy can provide a stable adaptive backbone, with performance depending on the instantiation. That backbone can then be paired with stronger domain-aware realizations when higher final precision is required.

\subsection{Interpretation}
Three observations emerge from the current experiments. First, the same resource loop remains workable across both domains without changing its organizing principle. Second, the framework is expressive enough that a richer realization can purchase better continuous refinement without changing the high-level logic. Third, route-based permutation search appears more tolerant to compact realization than multimodal continuous refinement in the current suite.

These observations should be interpreted cautiously. The experiments are intentionally small and diagnostic, not exhaustive. Their value lies in exposing a coherent behavior pattern rather than in claiming universal superiority.

The minimal realization preserves the closed-loop coupling among reward, local budget, communal reinvestment, role movement, and lifecycle turnover. What it intentionally sacrifices is high-resolution late-stage specialization: there is less elite precision refinement on the continuous side and less elaborate structure guidance on the discrete side. This is why the paper can support a positive framework claim while simultaneously documenting where compactness begins to cost performance.

\section{Discussion}
The results support the framework formulation and identify priorities for further work.

First, the framework seems strongest when one cares about \emph{how} adaptation is generated. MMAO offers a single explanatory language for search intensity, role change, reinvestment, and turnover. This is a methodological advantage over designs where each adaptive component is added independently.

Second, the continuous results show that compactness and final precision can be in tension. This is not a weakness of the framework idea itself, but it does imply that future high-performance instantiations will need better late-stage refinement while still keeping the resource loop central.

Third, the discrete results suggest that structure-aware local improvement and shared memory can coexist naturally with the metabolic controller. That makes combinatorial optimization a promising direction for future derivative MMAO studies.

Finally, the framework view clarifies a useful separation of concerns: the metabolic loop defines the family identity, whereas operators and proposal models define the domain-specific realization. This separation should make later derivative studies easier to organize, compare, and interpret.

Equally important, the present contraction study should not be judged by the same criterion as a strong benchmarking paper. Its main scientific role is to ask whether a stripped-down MMAO still behaves like MMAO. The answer supported here is yes: the compact version remains operational across both domains, and the richer version improves quality mainly by refining how the same loop spends, shares, and reuses search capital. The result identifies the effects of simplifying the resource-control mechanism.

\subsection{Limitations and Future Work}
Several limitations should be stated explicitly. First, the current empirical study is intentionally small and diagnostic; it is sufficient for framework-level interpretation, but not for strong competitiveness claims. Second, the continuous side currently shows that compact realization can sacrifice a nontrivial amount of late-stage precision, especially on multimodal landscapes. Third, the discrete evidence is limited to routing problems and should not yet be generalized to all combinatorial domains.

These limitations point directly to the next research steps. The first is to design stronger continuous proposal and refinement mechanisms that remain metabolically derived rather than externally attached. The second is to extend the empirical study to larger benchmark suites and richer statistical treatment. The third is to test whether the same framework logic transfers to other discrete structures such as assignment and scheduling. A longer-term direction is formal analysis of the communal-budget dynamics themselves, especially how normalized reward and role evolution interact to produce stable redistribution.

\section{Conclusion}
This paper presented MMAO as a resource-closed-loop framework for adaptive heuristic search. The framework is built around bounded private energy, a communal resource pool, normalized reward, continuous role adaptation, and resource-financed lifecycle turnover. A single control logic was instantiated in continuous and discrete search spaces and tested on a matched small-scale suite.

The results indicate that the framework is self-consistent and operational across domains. The continuous side shows that richer realization detail still matters for strong late-stage refinement, while the discrete side shows that the same metabolic controller can remain effective under a relatively lean route-based implementation. Overall, the study suggests that MMAO is best understood as a framework for generating adaptive search behavior from an endogenous resource economy rather than as a fixed operator set.

The minimal realization retains coupled resource control after reducing the mechanism set. The minimal realization does not prove that compact MMAO is always sufficient, but it does show that the family can be described by a small set of coupled resource principles rather than by an irreducible list of operators.

Future work can proceed in at least three directions: stronger precision-oriented continuous instantiations, broader statistically grounded empirical studies, and formal analysis of the budget dynamics that drive role and lifecycle adaptation. The formulation provides a basis for those studies.

\section*{Acknowledgement}
The related publications in Table~\ref{tab:mmao-research-series} provide complementary views of the MMAO research series.

\begin{table}[!ht]
\centering
\caption{Other publications in the MMAO research series.}
\label{tab:mmao-research-series}
\footnotesize
\setlength{\tabcolsep}{5pt}
\renewcommand{\arraystretch}{1.12}
\begin{tabular}{@{}p{\columnwidth}@{}}
\hline
\textbf{Related publication} \\
\hline
\emph{MMAO: A Metabolic Multi-Agent Optimizer with Endogenous Resource Allocation for Continuous and Discrete Optimization}~\cite{xu2026mmao}.\newline
\href{https://arxiv.org/abs/2606.28109}{arXiv:2606.28109}
\\[5pt]
\emph{A Large-Scale Empirical Evaluation of MMAO Under Fair-Budget Continuous and Discrete Benchmarks}~\cite{xu2026largescale}.\newline
\href{https://arxiv.org/abs/2606.31584}{arXiv:2606.31584}
\\[5pt]
\emph{MMAO-Dyn: A Metabolic Multi-Agent Optimizer for Dynamic Optimization}~\cite{xu2026mmaodyn}.\newline
\href{https://arxiv.org/abs/2607.00846}{arXiv:2607.00846}
\\[5pt]
\emph{MMAO-Cls: Metabolic Multi-Agent Optimization for Joint Feature Selection and Classifier Tuning}~\cite{xu2026mmaocls}.\newline
\href{https://arxiv.org/abs/2607.01539}{arXiv:2607.01539}
\\[5pt]
\emph{Mechanism and Stability Analysis of Metabolic Closed-Loop Metaheuristics}~\cite{xu2026mechanism}.\newline
\href{https://arxiv.org/abs/2607.01551}{arXiv:2607.01551}
\\[5pt]
\emph{MMAO: A Unified Resource-Regulated Controller for Cross-Domain Optimization}~\cite{xu2026unifiedmmao}.\newline
\href{https://doi.org/10.1109/ACCESS.2026.3752083}{IEEE Access, 2026}
\\[5pt]
\hline
\end{tabular}
\end{table}

\bibliographystyle{ieeetr}
\bibliography{Ref}

\end{document}